\documentclass[runningheads]{llncs}
\usepackage[T1]{fontenc}
\usepackage{graphicx,verbatim}
\usepackage{booktabs}
\usepackage{multirow}
\usepackage{graphicx}
\usepackage{colortbl}
\usepackage{xcolor}
\usepackage{amssymb}
\usepackage{amsmath}
\begin{document}
\title{Evidential Fusion Network for Multimodal Survival Prediction under Missing Modalities}
\titlerunning{Evidential Fusion under Missing Modalities}
%

\author{
Yucheng~Xing\inst{1,2}\thanks{Equal contribution.}
\and
Hailan~Mo\inst{1}\textsuperscript{\ensuremath{\star}}
\and
Zi~Wang\inst{3}
\and
Ling~Huang\inst{1,3}\textsuperscript{\ensuremath{\dagger}}
\and
Mengling~Feng\inst{1}
}


\authorrunning{Y. Xing et al.}

\institute{
National University of Singapore, Singapore
\and
National University of Singapore Guangzhou Research Translation and
Innovation Institute, Guangzhou, Guangdong, China
\and
Imperial College London, London, UK
}

\maketitle

\begingroup
\renewcommand{\thefootnote}{\ensuremath{\dagger}}
\footnotetext{Corresponding author: Ling Huang (\texttt{my@linghuang.co.uk}).}
\endgroup

\begin{abstract}
Recent multimodal survival prediction models have demonstrated strong predictive performance by leveraging complementary information across modalities. However, such models generally assume data completeness and exhibit limited robustness toward missing modalities, which are frequently encountered in real-world clinical settings.
We propose the Evidential Missing Modality Survival Fusion model for multimodal survival prediction under missing modalities. 
EMMS offers a straightforward, computationally effective approach to survival analysis without requiring a generative phase for missing data. By employing Dempster–Shafer theory and the newly introduced Gaussian Random Fuzzy Numbers for multimodal decision fusion, it considers both aleatoric and epistemic uncertainty alongside modality reliability for fusion. Moreover, the model treats missing modalities as vacuous evidence, preventing interference with available inputs and naturally reflecting increased uncertainty and calibrated predictions. Extensive experiments on four cancer datasets demonstrate state-of-the-art performance while providing calibrated and interpretable uncertainty estimates under incomplete multimodal observations, without extra computational overhead. Code is available at \url{https://github.com/YuchengXing99/EMMS}.

\keywords{Missing Modalities \and Multimodal learning \and Survival Analysis \and Evidential Fusion \and Gaussian Random Fuzzy Numbers \and Uncertainty}

\end{abstract}

\section{Introduction}
Accurate survival prediction is central to precision oncology, enabling personalized treatment planning and risk stratification. \cite{liu2026sparse,xing2026semantic,xing2026bridging} Although many state-of-the-art multimodal survival analysis methods, e.g., MCAT \cite{chen2021multimodal}, MOTCat \cite{xu2023multimodal}, EsurvFusion \cite{huang2025esurvfusion}, have provided more accurate predictions, the deployment of multimodal survival models in real-world clinical settings remains fundamentally constrained by missing modalities. In practice, a complete morpho–molecular profile that combines whole-slide images (WSIs) and transcriptomic measurements is often unavailable due to tissue exhaustion, sequencing costs, or heterogeneous digital pathology infrastructure across centers. 

Imputation-based approaches, such as SMIL \cite{ma2021smil} and M3Care \cite{zhang2022m3care}, reconstruct missing modalities through generative modeling or latent-space retrieval. Although effective in improving average performance, these methods risk hallucinating non-existent biological or morphological signals and often incur high computational cost, which is problematic in high-stakes clinical decision-making. In contrast, imputation-free approaches, such as EmbraceNet \cite{choi2019embracenet}, aim for graceful degradation by aggregating only available embeddings. However, most treat missingness as a masking or dropout issue, without explicitly quantifying the information loss induced by absent modalities. The above research also exposes a critical uncertainty gap in multimodal survival analysis \cite{huang2024evidential}. Clinical decision support requires distinguishing aleatoric uncertainty (intrinsic data noise) from epistemic uncertainty (lack of knowledge due to missing information), while most existing models provide only a “best guess,” without signaling their own ignorance.

To bridge this gap, we propose Evidential Missing Modality Survival Fusion (EMMS), a computationally efficient framework for multimodal survival prediction under missing modalities. EMMS is grounded in Dempster–Shafer (DS) evidential theory and represents each modality’s prediction as survival evidence parameterized by Gaussian Random Fuzzy Numbers (GRFNs) \cite{denoeux2023quantifying,xing2025dpsurv}. Different from existing methods, \textbf{we bypass the complexities of data reconstruction and heuristic degradation}; instead, we treat missing modalities as vacuous evidence by setting their epistemic strength to zero. This formulation leads to a closed-form, evidence-weighted aggregation where available modalities contribute proportionally to their reliability, while missing inputs exert no influence. Consequently, the total evidential strength serves as a built-in, interpretable metric of the epistemic uncertainty induced by incomplete data. By unifying interaction-aware multimodal representation learning with principled evidential fusion, EMMS produces calibrated survival predictions that degrade gracefully under missing input conditions. 

\section{Related Work}
\textbf{Multimodal Survival Analysis.}
The integration of digital pathology and transcriptomics has enabled survival prediction models that capture complementary morphological and molecular information \cite{ding2023pathology}. Recent methods emphasize interaction-based fusion. MCAT aligns genomic representations with discriminative WSI patches via co-attention \cite{chen2021multimodal}, while MOTCat enforces global structural consistency using optimal transport \cite{xu2023multimodal}. SurvPath further incorporates biological priors by representing transcriptomics as pathway tokens and modeling dense pathway–patch interactions with a multimodal Transformer \cite{jaume2024modeling}. Despite strong performance, these architectures assume complete modality availability, limiting robustness in clinical cohorts where WSI or RNA data are frequently missing.

\textbf{Learning with Missing Modalities.}
Handling missing modalities generally follows two primary paradigms: imputation-based or imputation-free learning. \textbf{Imputation-based} approaches reconstruct missing inputs through generative modeling or latent-space completion \cite{chen2020hgmf,chen2022m3ae,geng2022multimodal}. For example, SMIL employs prototype-weighted reconstruction with uncertainty regularization \cite{ma2021smil}, and M3Care retrieves task-relevant neighbors to compensate for absent modalities \cite{zhang2022m3care}. While improving average performance, these methods may hallucinate spurious biological or morphological signals, raising concerns in high-stakes clinical settings. \textbf{Imputation-free} methods instead aggregate only observed inputs \cite{gong2026embracing,yun2024flexmoemodelingarbitrarymodality}, for example, EmbraceNet introduces stochastic embracement for partial embeddings \cite{choi2019embracenet}. However, most treat missingness as masking or dropout, without explicitly modeling the epistemic uncertainty caused by absent evidence, often resulting in miscalibrated confidence.

\textbf{Evidential Survival Learning.}
Evidential learning framework based on Dempster–Shafer theory provides uncertainty-aware prediction by representing outputs as accumulated evidence \cite{denoeux2023parametric,denoeux2023reasoning,huang2024review,shafer1992dempster}. ENNreg first introduces Gaussian Random Fuzzy Numbers (GRFNs) for survival regression, separating aleatoric and epistemic uncertainty \cite{denoeux2023quantifying,huang2025evidential}. EsurvFusion extends this framework to multimodal fusion with learnable reliability discounting \cite{huang2025esurvfusion}. However, those models typically assume complete modalities. In this paper, we treat a missing modality as contributing vacuous evidence and naturally reducing total evidential strength in aggregations.

\section{Method}
\newcommand{\sigmah}{\sigma_h}

\subsection{Preliminaries}
\label{preliminaries}
\textbf{Gaussian Random Fuzzy Numbers.}
In epistemic random fuzzy sets theory, a GRFN is a random fuzzy subset of the real line defined by the membership function $\varphi(x; M, h) = \exp\!\Big(-\frac{h}{2} (x - M)^2\Big)$, whose mode $M$ is a Gaussian random variable with $M \sim \mathcal{N}(\mu, \sigma^2)$. A GRFN can be represented by $\tilde{Y} \sim \tilde{\mathcal{N}}(\mu, \sigma^2, h)$, where $\mu$ is the location parameter, and $\sigma$ and $h \in [0,+\infty)$ represent the aleatoric and epistemic uncertainties, respectively. 
Given a GRFN $\tilde{Y} \sim \tilde{\mathcal{N}}(\mu, \sigma^2, h)$, the degrees of plausibility and belief for any interval $[x, y]$ can be calculated using:
\begin{subequations}
\begin{align}
Pl_{\tilde Y}([x,y])
  &= \Phi\!\Big(\tfrac{y-\mu}{\sigma}\Big)
   - \Phi\!\Big(\tfrac{x-\mu}{\sigma}\Big)
   + pl_{\tilde Y}(x)\,
     \Phi\!\Big(\tfrac{x-\mu}{\sigmah}\Big) + pl_{\tilde Y}(y)\!
     \left[1-\Phi\!\Big(\tfrac{y-\mu}{\sigmah}\Big)\right],
     \label{eq:plxy}\\
Bel_{\tilde Y}([x,y])
  &= \Phi\!\Big(\tfrac{y-\mu}{\sigma}\Big)
   - \Phi\!\Big(\tfrac{x-\mu}{\sigma}\Big)
   - pl_{\tilde Y}(x)\!
     \Bigg[
       \Phi\!\Bigg(
         \tfrac{(x+y)/2-\mu+(y-x)h\sigma^{2}/2}{\sigmah}
       \Bigg)  \notag\\
  &\quad
       - \Phi\!\Big(\tfrac{x-\mu}{\sigmah}\Big)
     \Bigg] + pl_{\tilde Y}(y)\!
     \Bigg[
       \Phi\!\Bigg(
         \tfrac{(x+y)/2-\mu-(y-x)h\sigma^{2}/2}{\sigmah}
       \Bigg)
       - \Phi\!\Big(\tfrac{y-\mu}{\sigmah}\Big)
     \Bigg],
     \label{eq:belxy}
\end{align}
\label{eq:GRFN}
\end{subequations}
where $\sigma_h = \sigma\sqrt{1+h\sigma^{2}}$ and $pl_{\tilde Y}(x)= \frac{1}{\sqrt{1+h\sigma^{2}}}\exp\!\left(-\frac{h(x-\mu)^{2}}{2\,(1+h\sigma^{2})}\right)$ is the contour function and $\Phi$ denotes the standard normal cumulative distribution function.

\textbf{Vacuous Evidence.}
In the presence of missing modalities, the corresponding source provides no informative evidence. 
In the GRFN framework, this situation is naturally modeled by setting $h = 0$. In this case, the epistemic component vanishes and the GRFN degenerates into vacuous evidence, thereby representing complete ignorance.

\textbf{GRFNs for Survival Modeling.}
Let $T>0$ denote survival time and define $Y=\log T$ to map the positive domain onto the real line. 
Under a GRFN model $\tilde Y$, the survival function $S(t)=\mathbb{P}(T>t)$ is bounded as
\begin{equation}
Bel_{\tilde Y}((\log t,\infty))
\;\le\;
S(t)
\;\le\;
Pl_{\tilde Y}((\log t,\infty)).
\end{equation}

\begin{figure}[!t]
  \centering
  \includegraphics[width=1\textwidth]{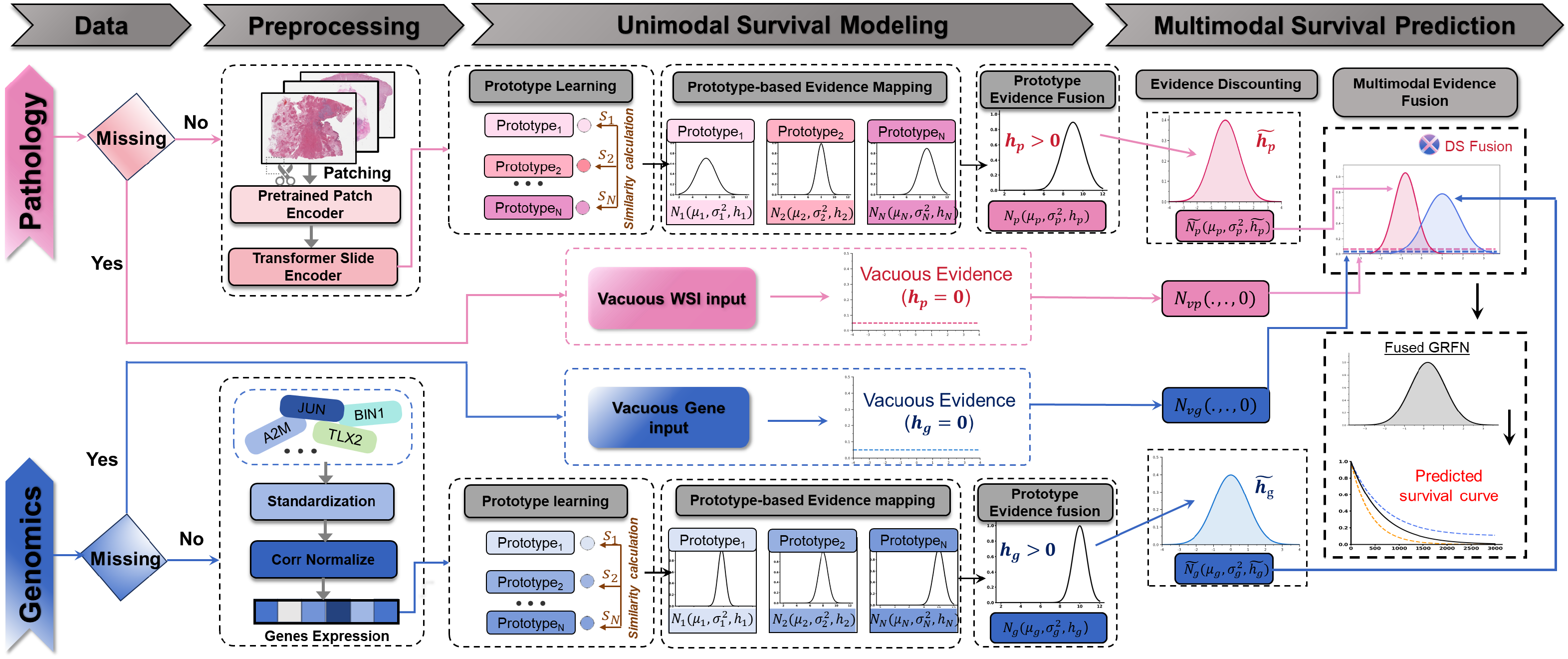}
  \caption{Overview of the proposed EMMS framework. WSI and gene modalities generate GRFN survival evidence, modeling both aleatoric and epistemic uncertainty. Missing modalities are treated as vacuous evidence and fused via Dempster–Shafer combination to produce calibrated survival predictions.
}
\end{figure}

\subsection{Unimodal Survival Modeling}
\label{single}
\textbf{Pathology prediction branch.}
Given a WSI, we tessellate it into $N_p$ patches $\{\mathbf{x}_{i,p}\}_{i=1}^{N_p}$ at 20x magnification. Each patch is encoded by a pretrained patch encoder $f_{\text{Patch}}$, yielding patch embeddings $\mathbf{h}_{i,p} = f_{\text{Patch}}\!\left(\mathbf{x}_{i,p}\right)$. The resulting set of patch features is then processed by the TITAN slide encoder \cite{ding2025multimodal}, which captures contextual dependencies among patches via a transformer-based architecture and aggregates them into a slide-level representation:
\begin{equation}
\mathbf{z}_p = f_{\text{TITAN}}\!\left(\{\mathbf{h}_{i,p}\}_{i=1}^{N_p}\right),
\end{equation}
where $\mathbf{z}_p\in\mathbb{R}^{D}$ serves as the global WSI feature for time-to-event modeling.

Following a prototype-based evidential time-to-event formulation \cite{huang2025evidential}, we map an input feature vector $\mathbf{z}_p$ to a survival evidence representation. Given $K$ prototypes $\{\mathbf{w}_k\}_{k=1}^{K}$, the prototype embedding layer evaluates the affinity between $\mathbf{z}_p$ and each prototype via a radial basis function kernel with a positive scale parameter $\gamma_k$:
\begin{equation}
s_k(\mathbf{z}_p)
=
\exp\!\left(
-\gamma_k^2
(\mathbf{z}_p-\mathbf{w}_k)^\top
\mathrm{Diag}(\mathbf{v}_p)\,
(\mathbf{z}_p-\mathbf{w}_k)
\right),
\end{equation}
where $\mathbf{v}_p \in \mathbb{R}_+^D$ is a learnable modality-specific scaling vector defining a diagonal Mahalanobis metric for adaptive dimension-wise reweighting.
For each prototype $\mathbf{w}_k$, the evidence mapping layer produces a GRFN
\begin{equation}
\tilde{Y}_k(\mathbf{z}_p) \sim \tilde{\mathcal{N}}\!\left(\mu_k(\mathbf{z}_p),\,\sigma_k^{2},\, s_k(\mathbf{z}_p)\,h_k\right),
\end{equation}
where $\sigma_k^{2}$ and $h_k$ denote the prototype-specific variance and precision, respectively. The location parameter is modeled linearly as
\begin{equation}
\mu_k(\mathbf{z}_p)=\beta_k^{\top}\mathbf{z}_p+\beta_{k0},
\end{equation}
with $\beta_k$ and $\beta_{k0}$ being learnable parameters.
Finally, evidence from all prototypes is integrated by the generalized product--intersection combination rule, yielding the aggregated GRFN
$\tilde{Y}_p(\mathbf{z}_p)\sim \tilde{\mathcal{N}}\!\left(\mu_p(\mathbf{z}_p),\,\sigma_p^{2}(\mathbf{z}_p),\,h_p(\mathbf{z}_p)\right)$, where
\begin{equation}
\begin{aligned}
\mu_p(\mathbf{z}_p) &= 
\frac{\sum_{k=1}^{K}s_k(\mathbf{z}_p)h_k\mu_k(\mathbf{z}_p)}
{\sum_{k=1}^{K}s_k(\mathbf{z}_p)h_k}, \quad
\sigma_p^2(\mathbf{z}_p) = 
\frac{\sum_{k=1}^{K}s_k^{2}(\mathbf{z}_p)h_k^{2}\sigma_k^{2}}
{\big(\sum_{k=1}^{K}s_k(\mathbf{z}_p)h_k\big)^2}, 
\end{aligned}
\end{equation}
and $h_p(\mathbf{z}_p) = \sum_{k=1}^{K}s_k(\mathbf{z}_p)h_k$. Here $\mu_p(\mathbf{z}_p)$ corresponds to the most plausible event-time estimate, $\sigma_p^{2}(\mathbf{z}_p)$ captures aleatoric uncertainty from data variability, and $h_p(\mathbf{z}_p)$ characterizes epistemic uncertainty due to model ignorance.

\textbf{Genomic prediction branch.}
For each individual, transcriptomic profile includes $N_g$ gene expression measurements and is represented as $\mathbf{z}_g = \{x_{i,g}\}_{i=1}^{N_g}$, where $x_{i,g} \in \mathbb{R}$. Following the same prediction framework as the pathology branch, the genomic branch produces survival evidence modeled as a GRFN
\[
\tilde{Y}_g(\mathbf{z}_g) \sim \tilde{\mathcal{N}}\!\big(\mu_g(\mathbf{z}_g),\,\sigma_g^{2}(\mathbf{z}_g),\,h_g(\mathbf{z}_g)\big).
\]

\subsection{Evidential Fusion under Missing Modalities}
\label{fusion} 
Let $\tilde{Y}_p(\mathbf{z}_p)$ and $\tilde{Y}_g(\mathbf{z}_g)$ denote the modality-specific survival evidence derived from pathology and genomics, respectively. We fuse these evidences using Dempster’s combination rule within the Dempster–Shafer framework, enabling principled aggregation of uncertainty.

To explicitly handle missing modalities, we introduce a binary availability indicator $r \in \{0,1\}$ for each modality. The corresponding GRFN is defined as 
\begin{equation}
\tilde{Y}^{(r)}(\mathbf{z}) =
\begin{cases}
\tilde{\mathcal{N}}\!\left(
\mu(\mathbf{z}),
\sigma^2(\mathbf{z}),
h(\mathbf{z})
\right), & r = 1, \\[10pt]
\tilde{\mathcal{N}}\!\left(
\cdot,\,\cdot,\,0
\right), & r = 0,
\end{cases}
\end{equation}
where $h(\mathbf{z})$ denotes the epistemic strength of evidence. Setting $h=0$ corresponds to vacuous evidence, representing complete ignorance in the DS sense. Importantly, this formulation does not approximate or reconstruct missing inputs; instead, it encodes their absence as zero evidential contribution.

Given two modalities with epistemic strengths $h_p$ and $h_g$, the fused GRFN
$\tilde{Y}_f \sim \tilde{\mathcal{N}}(\mu_f,\sigma_f^2,h_f)$
is obtained as
\begin{equation}
\mu_f
= \frac{d(h_p)\,\mu_p + d(h_g)\,\mu_g}{d(h_p) + d(h_g)},
\;
\sigma_f^{2}
= \frac{d(h_p)^{2}\sigma_p^{2} + d(h_g)^{2}\sigma_g^{2}}
       {(d(h_p) + d(h_g))^{2}},
\;
h_f
= d(h_p) + d(h_g),
\end{equation}
where $d(\cdot)$ denotes a monotonic discounting function defined as 
$d(h)=\rho \log(1+h)$, with $\rho \in (0,1)$ controlling the discounting intensity. 
The logarithmic form ensures bounded growth and diminishing sensitivity for large epistemic values, which stabilizes fusion under heterogeneous uncertainty levels.

The evidential fusion operation yields several desirable properties. First, fusion is evidence-weighted: modalities with higher epistemic strength contribute more to the fused prediction. Second, the total epistemic strength accumulates additively, providing an interpretable quantitative measure of overall confidence. Crucially, when a modality is missing ($r=0$), its epistemic parameter vanishes ($h=0$), and the fusion naturally reduces to the available modality:
$(\mu_f,\sigma_f^2,h_f)=(\mu_{available},\sigma_{available}^2,d(h_{available}))$.
Thus, missing modalities contribute vacuous evidence and do not interfere with observed data. Meanwhile, a reduction in total evidential strength explicitly reflects missingness-induced epistemic uncertainty.

\subsection{Loss function}
\label{sec:loss}

We follow the generalized negative log-likelihood for GRFN-based time-to-event learning \cite{huang2024evidential}.
The survival loss $\mathcal{L}_{\text{surv}}$ combines belief- and plausibility-based terms to handle both uncensored and right-censored samples:
uncensored cases supervise a small interval around $y=\log(t)$, while censored cases supervise the tail event $[y,\infty)$.

Let $\tilde{Y}_p$, $\tilde{Y}_g$ be the unimodal evidence and $\tilde{Y}_f$ the fused evidence.
With modality availability flags $r_p,r_g\in\{0,1\}$, our objective is
\begin{equation}
\mathcal{L}
=
r_p\,\mathcal{L}^{p}_{\text{surv}}
+
r_g\,\mathcal{L}^{g}_{\text{surv}}
+
\lambda_{\text{fus}}\,\mathbb{I}[r_p\lor r_g]\,
\mathcal{L}^{f}_{\text{surv}}
+
\lambda_{\text{align}}\,\mathbb{I}[r_p\land r_g]\,
\mathcal{L}_{\text{align}} ,
\end{equation}
where the complete-case alignment term is defined as
\begin{equation}
\mathcal{L}_{\text{align}}
=
\mathrm{KL}\!\left(\mathcal{N}(\mu_p,\sigma_p^2)\,\|\,\mathcal{N}(\mu_g,\sigma_g^2)\right)
+
\mathrm{KL}\!\left(\mathcal{N}(\mu_g,\sigma_g^2)\,\|\,\mathcal{N}(\mu_p,\sigma_p^2)\right).
\end{equation}
where $\mathrm{KL}(\cdot\|\cdot)$ denotes Kullback--Leibler divergence, and the symmetric KL between two Gaussians is used as a lightweight cross-modal consistency regularizer.

\section{Experiments}
\subsection{Experimental Setup}

\textbf{Datasets.}
Four cancer cohorts from The Cancer Genome Atlas (TCGA) are evaluated in this study, including Breast Invasive Carcinoma (BRCA), Stomach Adenocarcinoma (STAD), Kidney Renal Clear Cell Carcinoma (KIRC), and Lung Adenocarcinoma (LUAD). To prevent the model from exploiting center-specific artifacts such as staining patterns or scanner variations, we employ a 5-fold site-stratified cross-validation strategy \cite{howard2021impact}.

\textbf{Baselines.}
We employ DisPro \cite{Xu_2025_CVPR} and KNN \cite{pedregosa2011scikit} as an imputation-based baseline, and MUSE \cite{wu2024multimodal} and Flex-Moe \cite{yun2024flexmoemodelingarbitrarymodality} as an imputation-free baseline, and additionally consider MOTCat \cite{xu2023multimodal} in the no-missingness scenario. Following prior works \cite{ma2021smil,wang2023multi}, we simulate missing modalities by randomly masking fully paired samples under a total 60\% missing rate, yielding five pathology/genomics splits: (0\%,60\%), (20\%,40\%), (30\%,30\%), (40\%,20\%), and (60\%,0\%). For each missing setting, evaluation is conducted under pathology-only, genomics-only, and complete-modality settings. For EMMS, we set $K=50$ for each modality branch and initialize the prototypes deterministically using k-means clustering on the training features to promote diverse coverage of the feature space. Prototype precisions $h_k$ are initialized identically across branches, placing their epistemic strengths on a common scale. The fused survival loss jointly updates $h_p$ and $h_g$, preserving their cross-branch comparability during training.

\textbf{Evaluation Metrics.}
Model discrimination performance is assessed using the Concordance index (C-index) \cite{harrell1982evaluating}. Calibration performance is evaluated using the integrated Brier score (IBS) \cite{graf1999assessment}.
\begin{table}[t]
\centering
\tiny
\caption{Results under \textbf{0\% missing modality}.
$C$: C-index$\uparrow$, $S$: IBS$\downarrow$.
\textbf{G}, \textbf{P}, and \textbf{C} denote genomics-only, pathology-only,
and complete-modality evaluation. Best results are in \textbf{bold}, and our
method is highlighted in \colorbox{blue!5}{blue}.}

\label{tab:res_0}
\renewcommand{\arraystretch}{1.1}
\setlength{\tabcolsep}{0.6pt}
\begin{tabular}{c l | cc | cc | cc | cc}
\toprule
 & \multirow{2}{*}{\textbf{Method}} & \multicolumn{2}{c|}{\textbf{BRCA}} & \multicolumn{2}{c|}{\textbf{LUAD}} & \multicolumn{2}{c|}{\textbf{STAD}} & \multicolumn{2}{c}{\textbf{KIRC}} \\
\cmidrule(lr){3-4} \cmidrule(lr){5-6} \cmidrule(lr){7-8} \cmidrule(lr){9-10}
& & $C \uparrow$ & $S \downarrow$ & $C \uparrow$ & $S \downarrow$ & $C \uparrow$ & $S \downarrow$ & $C \uparrow$ & $S \downarrow$ \\
\midrule
\multirow{4}{*}{\rotatebox{90}{\textbf{G}}} 
& DisPro & 0.511±0.17 & 0.100±0.03 & 0.587±0.04 & 0.282±0.09 & 0.542±0.11 & 0.261±0.03 & 0.664±0.03 & 0.145±0.04 \\
& MUSE   & 0.661±0.10 & 0.134±0.07 & 0.586±0.03 & 0.282±0.03 & 0.472±0.09 & 0.299±0.04 & 0.664±0.11 & 0.168±0.07 \\
& Flex-MoE   & 0.582±0.08 & 0.068±0.04 & 0.579±0.05 & 0.150±0.04 & \textbf{0.598±0.06} & 0.172±0.06  & 0.552±0.12 & 0.104±0.03 \\
\rowcolor{blue!5} & Ours & \textbf{0.725±0.09} & \textbf{0.060±0.04} & \textbf{0.589±0.07} & \textbf{0.145±0.04} & 0.517±0.08 & \textbf{0.171±0.07} & \textbf{0.685±0.12} & \textbf{0.101±0.04} \\
\midrule
\multirow{4}{*}{\rotatebox{90}{\textbf{P}}} 
& DisPro & 0.641±0.19 & 0.104±0.06 & 0.600±0.11 & 0.179±0.05 & 0.539±0.09 & 0.206±0.06 & 0.639±0.15 & 0.150±0.05 \\
& MUSE   & 0.665±0.06 & 0.104±0.09 & 0.611±0.04 & 0.153±0.05 & 0.563±0.11 & 0.175±0.07 & \textbf{0.818±0.07} & 0.099±0.03 \\
& Flex-MoE   & 0.717±0.09 & 0.153±0.08 & 0.635±0.06 & 0.212±0.07 & 0.515±0.09 & 0.232±0.09  & 0.783±0.07 & 0.151±0.05 \\
\rowcolor{blue!5} & Ours & \textbf{0.720±0.10} & \textbf{0.057±0.03} & \textbf{0.638±0.08} & \textbf{0.139±0.04} & \textbf{0.586±0.10} & \textbf{0.170±0.07} & 0.801±0.06 & \textbf{0.093±0.03} \\
\midrule
\multirow{5}{*}{\rotatebox{90}{\textbf{C}}} 
& DisPro & 0.690±0.09 & 0.076±0.04 & 0.583±0.05 & 0.207±0.05 & 0.543±0.11 & 0.252±0.03 & 0.631±0.16 & 0.140±0.05 \\
& MUSE   & 0.682±0.09 & 0.099±0.07 & 0.601±0.03 & 0.165±0.04 & 0.488±0.09 & 0.209±0.05 & 0.732±0.10 & 0.111±0.05 \\
& MOTCat & 0.619±0.11 & 0.118±0.06 & 0.611±0.02 & 0.199±0.04 & 0.554±0.06 & 0.224±0.06 & 0.742±0.14 & 0.106±0.03 \\
& Flex-MoE & 0.723±0.07 & 0.066±0.04 & 0.612±0.07 & 0.160±0.03 & 0.499±0.09 & 0.201±0.07 & 0.768±0.05 & 0.100±0.03 \\
\rowcolor{blue!5} & Ours & \textbf{0.740±0.08} & \textbf{0.057±0.03} & \textbf{0.655±0.07} & \textbf{0.143±0.04} & \textbf{0.555±0.09} & \textbf{0.174±0.07} & \textbf{0.801±0.08} & \textbf{0.098±0.03} \\
\bottomrule
\end{tabular}
\end{table}
\begin{table}[t]
\centering
\tiny
\caption{Results under \textbf{60\% missing modality}.
$C$: C-index$\uparrow$, $S$: IBS$\downarrow$.
\textbf{G}, \textbf{P}, and \textbf{C} denote genomics-only, pathology-only,
and complete-modality test cases. Best results are in \textbf{bold}, and our
method is highlighted in \colorbox{blue!5}{blue}.}

\label{tab:res_60}
\renewcommand{\arraystretch}{1.1}
\setlength{\tabcolsep}{0.6pt}
\begin{tabular}{c l | cc | cc | cc | cc}
\toprule
 & \multirow{2}{*}{\textbf{Method}} & \multicolumn{2}{c|}{\textbf{BRCA}} & \multicolumn{2}{c|}{\textbf{LUAD}} & \multicolumn{2}{c|}{\textbf{STAD}} & \multicolumn{2}{c}{\textbf{KIRC}} \\
\cmidrule(lr){3-4} \cmidrule(lr){5-6} \cmidrule(lr){7-8} \cmidrule(lr){9-10}
& & $C \uparrow$ & $S \downarrow$ & $C \uparrow$ & $S \downarrow$ & $C \uparrow$ & $S \downarrow$ & $C \uparrow$ & $S \downarrow$ \\
\midrule
\multirow{5}{*}{\rotatebox{90}{\textbf{G}}} 
& KNN & 0.574±0.06 & 0.091±0.04 & 0.540±0.05 & 0.190±0.04 & 0.521±0.04 & 0.210±0.07 & 0.574±0.06 & 0.105±0.04 \\
& DisPro & 0.639±0.12 & 0.089±0.03 & 0.550±0.07 & 0.250±0.06 & 0.522±0.07 & 0.259±0.06 & 0.616±0.10 & 0.190±0.08 \\
& MUSE   & 0.650±0.08 & 0.114±0.09 & 0.589±0.06 & 0.176±0.03 & 0.514±0.12 & 0.229±0.07 & 0.607±0.18 & 0.137±0.07 \\
& Flex-MoE   & 0.608±0.07 & 0.108±0.09  & 0.573±0.02 & 0.165±0.01 & 0.524±0.04 & 0.189±0.02 & 0.624±0.06 & 0.104±0.02 \\
\rowcolor{blue!5} & Ours & \textbf{0.693±0.11} & \textbf{0.060±0.04} & \textbf{0.603±0.09} & \textbf{0.144±0.04} & \textbf{0.530±0.10} & \textbf{0.174±0.07} & \textbf{0.648±0.13} & \textbf{0.102±0.04} \\
\midrule
\multirow{5}{*}{\rotatebox{90}{\textbf{P}}} 
& KNN & 0.570±0.08 & 0.120±0.07 & 0.525±0.08 & 0.211±0.08 & 0.521±0.05 & 0.259±0.08 & 0.617±0.08 & 0.120±0.05 \\
& DisPro & 0.567±0.13 & 0.138±0.11 & 0.569±0.06 & 0.189±0.03 & 0.528±0.10 & 0.239±0.08 & 0.755±0.09 & 0.126±0.08 \\
& MUSE   & 0.647±0.10 & 0.142±0.13 & 0.609±0.07 & 0.180±0.03 & 0.567±0.08 & 0.216±0.06 & 0.779±0.07 & 0.104±0.03 \\
& Flex-MoE   & 0.594±0.05 & 0.083±0.04 & 0.595±0.01 & 0.167±0.04 & 0.476±0.02 & 0.226±0.05 & 0.753±0.02 & 0.103±0.01 \\
\rowcolor{blue!5} & Ours & \textbf{0.686±0.09} & \textbf{0.059±0.03} & \textbf{0.631±0.07} & \textbf{0.139±0.04} & \textbf{0.611±0.08} & \textbf{0.168±0.07} & \textbf{0.793±0.08} & \textbf{0.097±0.03} \\
\midrule
\multirow{5}{*}{\rotatebox{90}{\textbf{C}}} 
& KNN & 0.584±0.07 & 0.068±0.04 & 0.537±0.04 & 0.164±0.04 & 0.535±0.04 & 0.212±0.07 & 0.611±0.09 & 0.103±0.03 \\
& DisPro & 0.575±0.10 & 0.095±0.05 & 0.576±0.05 & 0.208±0.05 & 0.530±0.06 & 0.229±0.06 & 0.648±0.12 & 0.135±0.06 \\
& MUSE   & 0.698±0.08 & 0.105±0.09 & 0.624±0.05 & 0.154±0.03 & 0.541±0.09 & 0.191±0.05 & 0.737±0.13 & 0.104±0.04 \\
& Flex-MoE   & 0.605±0.08 & 0.070±0.01 & 0.606±0.01 & 0.161±0.01 & 0.481±0.04 & 0.196±0.01 & 0.713±0.03 & 0.103±0.02 \\
\rowcolor{blue!5} & Ours & \textbf{0.715±0.06} & \textbf{0.059±0.04} & \textbf{0.637±0.07} & \textbf{0.142±0.04} & \textbf{0.576±0.11} & \textbf{0.174±0.07} & \textbf{0.753±0.10} & \textbf{0.102±0.04} \\
\bottomrule
\end{tabular}
\end{table}

\begin{figure}[!htbp]
  \centering
  \includegraphics[width=0.7\textwidth]{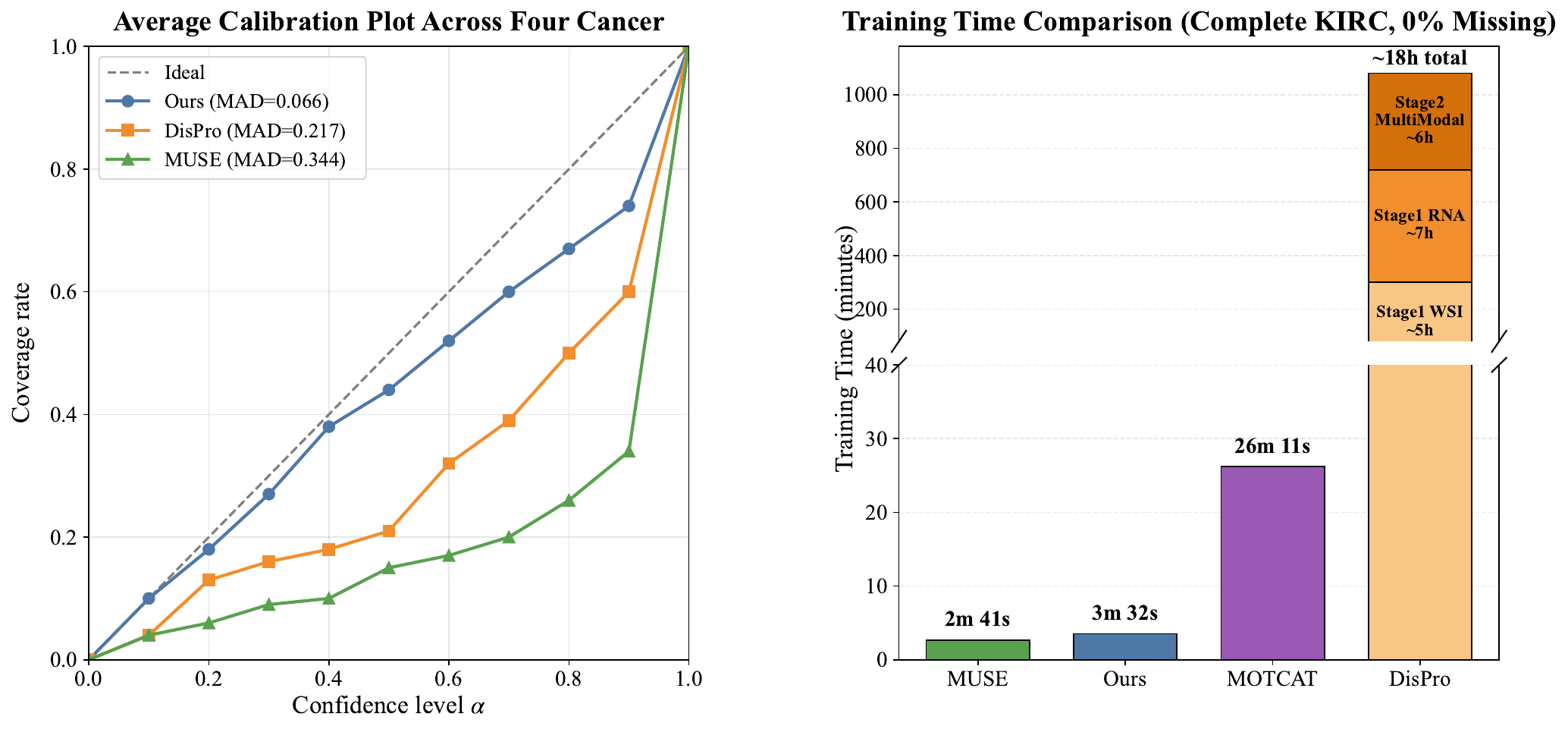}
  \caption{Left: calibration curves (MAD, mean absolute deviation from the diagonal used to quantify calibration). Right: training time comparison across methods.}
  \label{fig:calibration}
\end{figure}

\subsection{Performance comparison under different missing scenarios}
Table~\ref{tab:res_0} reports the results under the 0\% missing-modality setting.
Across the four datasets, our method achieves the best performance in most cases.
Compared with the baselines, it generally attains higher C-index and lower IBS, 
indicating improved discrimination and calibration under fully observed conditions. Table~\ref{tab:res_60} presents the results under the 60\% missing-modality setting.
Across the four datasets, our method achieves the best performance in all cases and demonstrates strong robustness under substantial modality absence. Overall, these results demonstrate the effectiveness and robustness of our framework across both complete and incomplete modality scenarios.

\subsection{Calibration Analysis and Computational Cost}
Figure~\ref{fig:calibration} shows that our method remains closer to the diagonal line across all four datasets, indicating better calibration than the compared baselines. This consistent behavior suggests improved confidence estimation and highlights the reliability of our uncertainty quantification in survival prediction. In terms of computational cost, our approach is comparable to the imputation-free baseline MUSE, while remaining substantially more efficient than the imputation-based method DisPro, indicating that EMMS achieves strong robustness and calibration performance without introducing additional computational overhead.

\section{Conclusion}
In this paper, we introduced EMMS, a survival analysis framework for incomplete multimodal data. Leveraging Dempster–Shafer theory and GRFNs, it models missing modalities as vacuous evidence without introducing extra computational cost, preventing fusion contamination while capturing increased predictive uncertainty under missingness. Experiments on four public cancer datasets show SOTA discrimination and calibration across diverse clinical cohorts and varying missing-modality patterns, standing as a scalable, computation-effective, uncertainty-aware, and theoretically grounded solution for personalized medicine in realistic clinical settings.

\begin{credits}
\subsubsection{\ackname}
This research was supported by the National Medical Research Council (NMRC) Healthy and Meaningful Longevity--Cognition Grant (NICCOG2024-0028), the AI for Public Health Program at the Saw Swee Hock School of Public Health, National University of Singapore, and the Dame Julia Higgins Postdoc Collaborative Research Fund. This work was also supported by the NUS Guangzhou Research Translation and Innovation Institute (GRTII)-Affiliated PhD Scholarship Program.

\subsubsection{\discintname}
The authors have no competing interests to declare that are relevant to the content of this article.
\end{credits}

\bibliographystyle{splncs04}
\bibliography{reference}

\end{document}